\documentclass[runningheads]{llncs}
\usepackage{graphicx}
\newcommand{\nunito}{}

\newcommand{\figlabel}[1]{\label{Fi:#1}}

\def\Figref#1{Figure~\ref{Fi:#1}}
\def\twoFigref#1#2{Figures~\ref{Fi:#1} and \ref{Fi:#2}}

\usepackage[normalem]{ulem}

\usepackage{tikz}

\usetikzlibrary{mindmap,calc,trees,arrows,backgrounds}

\tikzset{label_small/.style={font=\scriptsize}}
\tikzset{robustness_large/.style={font=\Large\bfseries}}
\tikzset{
    between/.style args={#1 and #2}{
         at = ($(#1)!0.5!(#2)$)
    }
}

\definecolor{lforange}{HTML}{FF9E1B}
\definecolor{lfblue}{HTML}{418FDE}
\definecolor{lfdarkblue}{HTML}{051C2C}
\definecolor{lforangesoda}{HTML}{F55D3E}
\definecolor{lfyellowred}{HTML}{FFC857}
\definecolor{lfraisinblack}{HTML}{212738}

\begin{document}
\title{Robustness testing of AI systems: A case study for traffic sign recognition}
\titlerunning{Robustness testing of AI systems}
%
\author{Christian Berghoff\inst{1}\thanks{Authors are listed in alphabetical order.} \and
Pavol Bielik\inst{2} \and
Matthias Neu\inst{1} \and
Petar Tsankov\inst{2} \and
Arndt von Twickel\inst{1}
}
\authorrunning{C. Berghoff et al.}
%
\institute{Federal Office for Information Security, Bonn, Germany\\ \email{<first name>.<last name>@bsi.bund.de} \and LatticeFlow, Zurich, Switzerland\\ \email{<first name>@latticeflow.ai}} 
%
\maketitle              
\begin{abstract}
In the last years, AI systems, in particular neural networks, have seen a tremendous increase in performance, and they are now used in a broad range of applications. Unlike classical symbolic AI systems, neural networks are trained using large data sets and their inner structure containing possibly billions of parameters does not lend itself to human interpretation. As a consequence, it is so far not feasible to provide broad guarantees for the correct behaviour of neural networks during operation if they process input data that significantly differ from those seen during training. However, many applications of AI systems are security- or safety-critical, and hence require obtaining statements on the robustness of the systems when facing unexpected events, whether they occur naturally or are induced by an attacker in a targeted way. As a step towards developing robust AI systems for such applications, this paper presents how the robustness of AI systems can be practically examined and which methods and metrics can be used to do so. The robustness testing methodology is described and analysed for the example use case of traffic sign recognition in autonomous driving.

\end{abstract}
\section{Introduction}

AI systems based on neural networks have tremendously increased their performance over the course of the last years and are now used in a plethora of very diverse areas. The improved performance is in particular based on a strong increase in available computing power and data, but also on theoretic advances in the area of AI in general and in the design of the networks in particular. Current applications of neural networks range from predictions based on structured data to natural language processing and computer vision tasks. In the latter domain, AI systems are for instance used in biometrics for identifying people, and as an important building block of (partially) autonomous driving for processing and analysing sensor data.\\

However, unlike the classical systems of symbolic AI, neural networks can, in most cases, not be manually developed by experts, but are instead trained on the basis of large data sets so as to provide the desired functionality. As a result of this training procedure, several problems arise when evaluating the behaviour of AI systems after they have been deployed to live operation. On the one hand, the systems are strongly dependent on the quality and representativeness of training data, whose amount is necessarily limited and which, in most realistic application scenarios, cannot cover all possible inputs \cite{Berghoff2020}. 
Furthermore, it is becoming increasingly evident that even if we could easily obtain much larger data sets, this alone will not be sufficient to ensure the safe operation of these models~\cite{Geirhos2020a,damour2020underspecification}.\\

As a result, significant effort has been recently devoted to developing techniques capable of providing formal robustness guarantees~\cite{Huang2017,Katz2017,Gehr2018,Singh2019}.
Such guarantees are of paramount importance, especially when considering use cases where severe damages may potentially arise. As an example, malfunction of an AI system used in autonomous driving can lead to significant material loss and in extreme cases to fatalities.
Unfortunately, such techniques are currently not applicable to state-of-the-art computer vision models due to their limited scalability and the limited type of robustness properties they support.\\

This article focuses on the complementary approach of developing a methodology for testing the robustness of AI systems. This allows the developers of the AI systems to assess the robustness of any state-of-the-art system in a principled way and to identify failure modes that need to be explored in more detail and remedied before the system can be deployed. While this empirical approach cannot provide absolute guarantees, the amount of robustness testing carried out can be adjusted to achieve the required confidence level about the model correctness. It allows systematically taking into account prior human knowledge by defining task-specific robustness properties, and thus aids in making the development of robust neural networks more efficient. In what follows, we will briefly describe the main components of our work and instantiate the approach to the concrete use case of traffic sign recognition. 

\section{Related Work}
The robustness of AI systems has been extensively studied in the literature, but the large majority of research has focused on their robustness to attacks specially designed to break the system, an area commonly referred to as adversarial machine learning (AML). Starting with the first publication \cite{Szegedy2014} applying adversarial attacks (more generally known as evasion attacks years in advance \cite{Dalvi2004}) to neural networks, many approaches to mitigating the vulnerability of neural networks to these attacks have been developed and many of them have been broken subsequently \cite{Biggio2018}. Another similar, albeit somewhat less popular subject of research are poisoning attacks \cite{Biggio2018}. One common feature of research in AML is the lack of a standardised and agreed-upon framework for evaluating robustness, although some progress has been made towards this end \cite{Carlini2019}.\\

When considering research on the robustness of AI systems to perturbations that may occur naturally during operation and should thus be much easier to cope with than malicious attacks, results are much more scarce. Some publications deal with the robustness to natural perturbations and propose measures for increasing it \cite{Michaelis2019,Ponn2020,Kim2021}, but they do not set out to perform a fine-grained and systematic assessment of the phenomenon. Probably the work most closely related to ours is \cite{Temel2017_NIPSW,temel2019traffic}. Similar to our work, the authors define a set of robustness properties and assess the model robustness against them. However, while the main goal of their work is to introduce a new benchmark, the goal of our work is a thorough assessment of state-of-the-art models. As a result, (i) we assess the robustness of significantly better models that achieve 99\% standard accuracy (compared to models with at most 90\% accuracy used in \cite{Temel2017_NIPSW}), (ii) we show the need of considering robustness properties jointly, rather than in isolation, and (iii) we focus on an iterative methodology to assess and improve model robustness (which includes eliciting new properties and discovering model failure modes), rather than proposing a fixed data set.\\

One line of work that can yield results for this research question is that of formal verification \cite{Huang2017,Katz2017,Gehr2018,Singh2019}. The main limitations of formal verification are that the methods developed so far do not scale to large neural networks used in practice and that the type of robustness properties that can be efficiently encoded is limited. For example, the majority of the current works consider only norm-constrained pixel perturbations, and verifying even simple geometric transformations such as rotations is highly non-trivial~\cite{geometric_certification}. Further, formal verification approaches suffer from the issue of incomplete specification, since the number of robustness properties they can encode is limited (mostly pixel perturbations). As a result, one has to be careful when interpreting their results and make sure not to incur a false sense of security by performing a thorough assessment of a wide range of the robustness properties, not only a subset of those for which formal verification can provide strong guarantees.

\begin{figure}
\includegraphics[width=\textwidth]{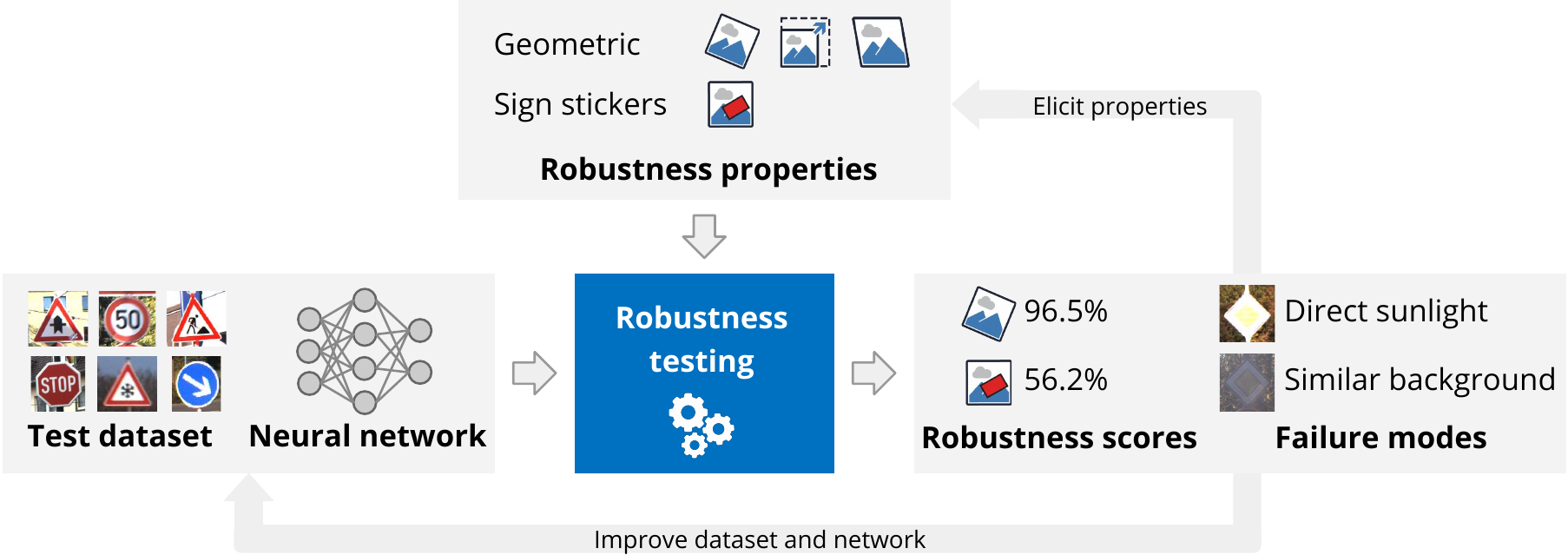}
\caption{Robustness assessment of neural networks. The testing approach takes as input a neural network and its test data set, along with robustness properties that capture important environmental conditions (e.g., camera position, sign stickers), and assesses the neural network against these together with its common failure modes.}
\figlabel{intro}
\end{figure}

\section{Approach}\label{sec:approach}
This section presents details about our case study on the robustness testing of traffic sign classifiers. \Figref{intro} provides an overview of the approach. Robustness testing is performed on a neural network based on a test data set and a specification of certain properties. A metric is used to compute the robustness scores. Apart from identifying common failure modes of the model, the results can be used to understand how to enhance the data set and to define additional custom properties, which improve both the neural network and the testing process. More detailed information on the individual components for the case study on traffic sign classifiers is described in sections \ref{sec:data}--\ref{sec:metric}. The general methodology depicted in \Figref{intro} can be applied to other use cases by exchanging and adjusting the components in a suitable way.

\subsection{Models}\label{sec:models}
The following models, which were trained by the authors, were used for the robustness testing: 
\begin{itemize}
 \item The first neural network, called pre-trained, has 99.0\% accuracy. It uses the pre-trained Inception-v3 model \cite{Szegedy2016}, which is trained on the ImageNet data set \cite{Deng2009} and fine-tuned for the GTSRB data set~\cite{Houben2013}.
\item The second one, called self-trained, has 97.4\% accuracy. It uses an architecture based on Inception-v3 but with reduced size and without pre-training.
\end{itemize}

Both networks were trained on the GTSRB data set with a data augmentation policy that applies the following random transformations: random cropping of a portion of the original image with a side length between 0.6 and 1.0 of its original length, rotation by an angle within $-15$ and $+15$ degrees, colour changes -- brightness, contrast, saturation, and hue -- by a factor of up to 0.1, and random change to a gray-scale format with a probability of 0.1. The transformed images are then scaled to the neural networks' expected resolution -- $32 \times 32$ pixels for the self-trained network and $299 \times 299$ pixels for the pre-trained network.

\subsection{Data Set}\label{sec:data}
We use the German Traffic Sign Recognition Benchmark data set (GTSRB) \cite{Houben2013}, as illustrated in \Figref{gtsrb}. This data set consists of 39,209 colour images used for training and 12,630 images for testing, each assigned to one of 43 classes. The data set images have different resolutions, with height and width ranging between 25--266 and 25--232 pixels, respectively. 

\begin{figure}[h]
\centering
\begin{tikzpicture}
\foreach \i in {0,...,42}
{
\pgfmathtruncatemacro{\x}{mod(\i, 15)};
\pgfmathtruncatemacro{\y}{-1 * (\i / 15)};
\node[inner sep=0pt] at (0.82*\x, 0.82*\y) 
    {\includegraphics[height=.05\textwidth, width=.05\textwidth]{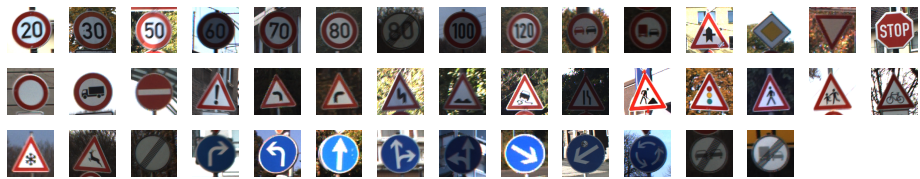}};
}
\end{tikzpicture}
\vspace{-0.5em}
\caption{Example traffic signs drawn from the GTSRB data set~\cite{Houben2013}, one for each class.}
\figlabel{gtsrb}
\end{figure}

\subsection{Robustness properties}\label{sec:perturbations}
The robustness properties used were chosen according to the ambient conditions of the use case. Traffic sign recognition, and autonomous driving in particular, is a very challenging but also relevant use case in this respect. Since autonomous driving technologies are deployed in the real world and on the roads, ambient conditions can change dramatically depending on the time of day, season or weather. Traffic signs may also to some extent be rotated, occluded and damaged. The sensors used to capture images of the traffic signs may themselves be degraded from long use, damaged or soiled, and the quality of the images captured can strongly vary depending on the specific lighting conditions. This situation is in (stark) contrast to other sensitive use cases such as medical image classification or biometrics used in border control, where ambient conditions can be standardised to a (much) larger extent, and the conditions after deployment can be made reasonably close to the ones considered during training.\\

The following list provides a brief overview of the robustness properties used. For the complete list of properties along with the used robustness bounds, we refer the reader to our detailed technical report~\cite{bericht2020}. 
\begin{itemize}
 \item \textbf{Image noise} includes Gaussian noise, uniform noise and impulse noise, each of them modelling different deviations from optimal conditions in image capture or pre-processing.
 \item \textbf{Pixel perturbations} define the robustness over individual pixels, each of which is allowed to change independently from other pixels subject to a total perturbation budget measured in the $L_0$ or $L_\infty$ norm. The $L_0$ norm allows modifying a limited amount of pixels to any extent, while the $L_\infty$ norm allows changing all pixels by less than a certain threshold.
 \item \textbf{Geometric transformations} model different orientations or positions of the traffic signs as well as faults in image acquisition and pre-processing. The perturbations considered are rotation, translation, scaling, shearing, blurring, sharpening and flipping (flipping was only considered on images whose class was not changed by applying it).
 \item \textbf{Colour transformations} model changes in lighting conditions or colour post-processing performed on the images. The colour transformations considered are brightness, contrast, saturation, hue, gray-scale and colour-depth.
\end{itemize}

\subsection{Metric}\label{sec:metric}
The metric used to assess the robustness of a model to a specified property is the robustness score, defined as the fraction of robust samples in the data set. A~sample is called robust to a given property if the network produces the correct label for all transformations captured by the property. For example, suppose the model achieves 86.5\% robustness to rotations up to 15 degrees.
This indicates that for 13.5\% of the input samples, a rotation of the original image within 15 degrees was found which causes the model to predict the wrong label. Robustness is defined only over input images for which the model already predicts the correct label as all incorrect inputs are trivially non-robust.

\begin{figure}[t]
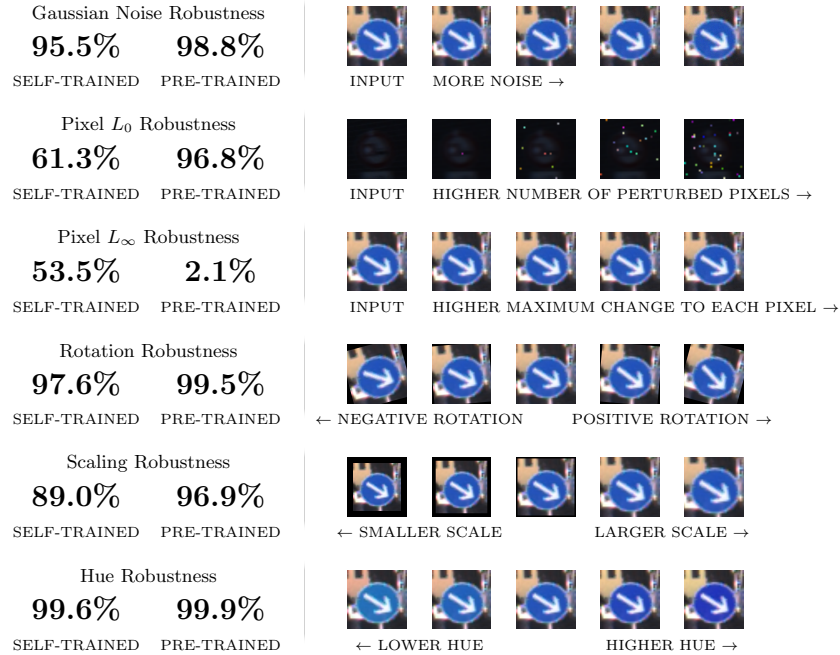

\scalebox{.8}{\hspace{4.5em}\input{figures/image_noise_overview}}
\scalebox{.8}{\hspace{4.5em}\input{figures/pixel_l0_overview}}
\scalebox{.8}{\hspace{4.5em}\input{figures/pixel_linf_overview}}
\scalebox{.8}{\hspace{4.5em}\input{figures/rotation_overview}}
\scalebox{.8}{\hspace{4.5em}\input{figures/scaling_overview}}
\scalebox{.8}{\hspace{4.5em}\input{figures/hue_property}}
\vspace{-0.6em}
\caption{Overview of the models' robustness to different transformations.}
\figlabel{basic_robustness}
\end{figure}

\section{Results}\label{sec:results}

This section provides a summary of the results obtained using the data set, models, robustness properties and metric from section \ref{sec:approach}. For comprehensive~results, as well as exact definitions of the tested properties, the reader is referred to \cite{bericht2020}.

\subsection{Basic Robustness Tests}\label{sec:basic_tests}
\Figref{basic_robustness} shows the robustness of the models with respect to the robustness properties as outlined in section \ref{sec:perturbations}. 
Each part of \Figref{basic_robustness} is organised as follows: The robustness score of both the self-trained and pre-trained model to the respective property is given, and the result of applying different intensities of the property on the original image is provided to show the visual impact for easy human inspection. \Figref{basic_robustness} shows that the robustness scores significantly differ, both between models and properties. Whereas both models are relatively robust (score $\geq 93.9\%$) to Gaussian noise, rotations and colour transformations such as changes in brightness (\Figref{adaptive_brightness}) and hue, the robustness of the self-trained model to rescaling the input is lower and its robustness to pixel $L_0$ perturbations even much lower, with more than one-third of the inputs being non-robust. Of the properties selected for \Figref{basic_robustness}, the robustness to pixel $L_\infty$ perturbations exhibits by far the lowest scores, with only about half the inputs being robust for the self-trained model and virtually none for the pre-trained one.

\begin{figure}[t]
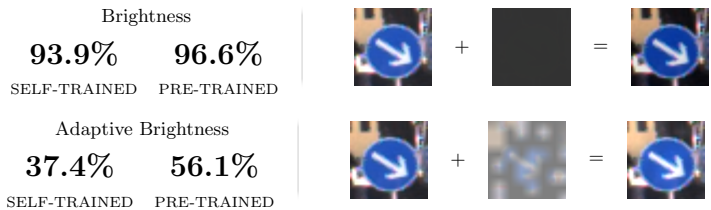

\centering
\scalebox{.8}{\input{figures/brightness_basic}}
\vspace{-0.15em}
\scalebox{.8}{\input{figures/brightness_adaptive}}
\vspace{-1em}
\caption{Robustness comparison of the basic (top) and adaptive brightness (bottom). Basic brightness applies the same transformation to the whole image, whereas adaptive brightness allows different transformations for different parts of the image.}
\figlabel{adaptive_brightness}
\end{figure}

\begin{figure}[t]
\centering
\scalebox{.8}{\input{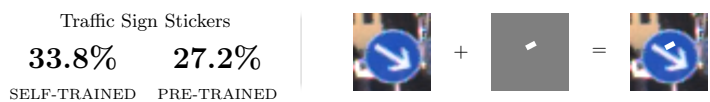}}
\vspace{-1em}
\caption{Task-specific property that inserts a single sticker of varying position, size and orientation on the traffic sign.}
\figlabel{stickers}
\end{figure}

\subsection{Stronger and Task-specific Properties}\label{sec:task-specific}
In addition to the generic and basic computer vision properties considered in section \ref{sec:basic_tests}, many of which naturally transfer to other use cases and application domains, the robustness of the models for traffic sign recognition was assessed with respect to more-task specific and stronger properties. Generally speaking, these properties better reflect specific ambient conditions that may occur in this use case. \twoFigref{adaptive_brightness}{stickers} show the results for two task-specific properties. On the one hand, the adaptive brightness property is defined as a generalisation of the basic brightness transformation. While the basic brightness transformation applies brightness changes uniformly to the whole image, the adaptive brightness transformation makes fine-grained changes, allowing some parts of the image to become brighter, and others darker. This better reproduces situations that can occur in practice, as a result of the material properties of object surfaces with respect to absorption and reflection of light. On the other hand, the robustness to generic stickers placed on the traffic signs is assessed. Especially in metropolitan areas, it is not uncommon for such stickers to be placed on traffic signs. As can be seen from \twoFigref{adaptive_brightness}{stickers}, the models become much more brittle when generalising the basic brightness transformation to its adaptive version, and their robustness to stickers is very low indeed.


\subsection{Testing Combinations}\label{sec:combinations}
So far, the robustness properties were analysed independently from each other. However, a combination of all of the properties with different strengths reflects real-world conditions more realistically. As a first step towards evaluating these situations, the properties can be composed together (e.g., combining brightness with rotation), effectively creating a new set of properties with additional degrees of freedom that should be assessed for robustness. \Figref{combined_overview} provides an overview of robustness scores of the models when tested against combinations of two properties. In most cases, the robustness scores for combinations decrease. For instance, combining translation and scaling yields a robustness score of 85.5\%, whereas the robustness to translation and scaling only is 94.7\% and 93.0\%, respectively. The combination thus decreases robustness by an additional 7.5\% as compared to the lower of the two individual robustness scores.

\input{figures/combined_overview}

\section{Discussion}
The results in section \ref{sec:results} show that significant differences in robustness exist both between models and properties. Variations in robustness scores can be partly ascribed to differences in the structure of the models and in the resolution of their input images (which were taken into account in testing in order not to distort the results). The different levels of brittleness the models exhibit to different properties at least partly correlate with the degree of representation of these properties in the training data set or the data augmentation process. It is assumed that, hereby, the models are actively endowed with a certain baseline robustness against such properties. It is concerning to see that the models' robustness to some properties is very low indeed, and using such brittle models in safety- and security-relevant use cases seems audacious.\\

The more detailed analyses of stronger and task-specific properties as well as of combinations of properties outlined in sections \ref{sec:task-specific} and \ref{sec:combinations} show that performing a standard set of basic robustness tests provides a good baseline, but more comprehensive tests can yield higher levels of confidence and better insights as to under what conditions the models are robust.\\

As an additional refinement, the aggregate robustness scores can be replaced with more precise plots of model robustness across the perturbation parameter space (e.g., \cite{engstrom19a,fawzi18geometry}). This can provide detailed insights into the specific regions where models may fail, which is especially beneficial in safety- and security-critical applications. 
Beyond that, a careful, fine-grained inspection of the robustness results can help pinpoint failure modes, i.e. specific conditions where model robustness is lower than in others (e.g., direct sunlight or very low brightness, as shown in \Figref{failure_modes}). 
Improving the training data sets based on the results obtained in the robustness tests, e.g. by using specially selected, augmented or synthesised data, seems a promising direction for alleviating their shortcomings.\\ 

\input{figures/failure_modes_overview}

The results presented in section \ref{sec:results} are those obtained using the specific use case of traffic sign recognition, a specific data set and two specific models trained with very limited resources. It is to be expected that models deployed in the real-world by car vendors will have been trained using much larger resources and will, therefore, exhibit higher robustness on average. Nevertheless, a principled procedure for testing their robustness is necessary and the approach discussed in this article yields one possible solution.

\section{Conclusion and Outlook}

The tests that were performed in the scope of this work mainly focus on the robustness of models against naturally occurring, random properties. They only address to a limited extent the stability of the models with respect to perturbations crafted by an attacker in a targeted way. On the one hand, it is to be expected that the models' robustness can be further decreased by using specialised attacks and that remedying this problem is much harder than improving robustness to certain naturally occurring failure modes as identified by the tests. On the other hand, the effort required for performing different kinds of attacks strongly varies and, therefore, not all attacks are realistic depending on the use case, the motivation and the capabilities of possible attackers.\\

The testing methodology was applied to the special use case of traffic sign recognition. The results show that performing such tests in a principled way is feasible, and they yield valuable insights into the existing limitations of the models. As a next step, such tests should also be applied to other use cases, in order to obtain the respective results. In doing so, the questions to what extent the testing methodology from traffic sign recognition can be transferred to other use cases and where it needs to be adapted should be addressed. Whereas the basic tests do not have any specific connection to the example use case considered and can probably be transferred easily to check models solving different computer vision problems, the task-specific properties may well need to be adjusted to reflect the specific ambient conditions that may arise in such new use cases. More generally, the abstract testing methodology might also be generalised to AI models solving problems not based on image data, but on data from other input domains, e.g. acoustic signals or even structured tabular data, and to other model architectures beyond neural networks.\\

For AI models to be used in safety- and security-critical areas such as (partly) autonomous driving in the years to come, a standardised methodology and concrete test criteria will be required in order to assess and evaluate the robustness of these models with respect to random as well as targeted perturbations. For ensuring an adequate level of safety and security, such criteria must be developed, and checking compliance with them must be made compulsory.\\

Another question that should be further studied is under what conditions and to what extent it is feasible not only to test the robustness of the models in an empirical, though principled, way but also to obtain formally verified statements, which could guarantee the required level of safety and security. Intense research should be devoted to this question, while aiming at practical applicability of solutions at least as a mid-term goal. Combining robustified models with a reject option for specific, non-verifiable regions of the input space might be a first step in this direction. In addition to that, further research effort should be devoted to investigating defensive measures that are able to ensure a high security level even when facing strong attackers, i.e. adaptive attackers with knowledge of the security measures in place (e.g., \cite{adaptive20tramer}), and to developing fundamental approaches for increasing the robustness of AI models.

\section{Acknowledgements}
The authors would like to thank the reviewers for their helpful suggestions.

%
%
%
 \bibliographystyle{splncs04}
 \bibliography{paper_robustness}

\begin{thebibliography}{10}
\providecommand{\url}[1]{\texttt{#1}}
\providecommand{\urlprefix}{URL }
\providecommand{\doi}[1]{https://doi.org/#1}

\bibitem{geometric_certification}
Balunovic, M., Baader, M., Singh, G., Gehr, T., Vechev, M.: Certifying
  geometric robustness of neural networks. In: Wallach, H., Larochelle, H.,
  Beygelzimer, A., d'Alch\'{e} Buc, F., Fox, E., Garnett, R. (eds.) Advances in
  Neural Information Processing Systems. vol.~32. Curran Associates, Inc.
  (2019)

\bibitem{Berghoff2020}
Berghoff, C., Neu, M., von Twickel, A.: {Vulnerabilities of Connectionist {AI}
  Applications: Evaluation and Defense}. Frontiers Big Data  \textbf{3}, ~23
  (2020). \doi{10.3389/fdata.2020.00023}

\bibitem{bericht2020}
Bielik, P., Tsankov, P., Krause, A., Vechev, M.: {Reliability Assessment of
  Traffic Sign Classifiers}. Tech. rep., {Bundesamt f\"ur Sicherheit in der
  Informationstechnik} (2020), https://www.bsi.bund.de/ki

\bibitem{Biggio2018}
Biggio, B., Roli, F.: Wild patterns: Ten years after the rise of adversarial
  machine learning. Pattern Recognit.  \textbf{84},  317--331 (2018).
  \doi{10.1016/j.patcog.2018.07.023}

\bibitem{Carlini2019}
Carlini, N., Athalye, A., Papernot, N., Brendel, W., Rauber, J., Tsipras, D.,
  Goodfellow, I.J., Madry, A., Kurakin, A.: On evaluating adversarial
  robustness. CoRR  \textbf{abs/1902.06705} (2019)

\bibitem{Dalvi2004}
Dalvi, N.N., Domingos, P.M., Mausam, Sanghai, S.K., Verma, D.: Adversarial
  classification. In: Kim, W., Kohavi, R., Gehrke, J., DuMouchel, W. (eds.)
  Proceedings of the Tenth {ACM} {SIGKDD} International Conference on Knowledge
  Discovery and Data Mining. pp. 99--108. {ACM} (2004).
  \doi{10.1145/1014052.1014066}

\bibitem{damour2020underspecification}
D'Amour, A., et~al.: Underspecification presents challenges for credibility in
  modern machine learning. CoRR  \textbf{abs/2011.03395} (2020)

\bibitem{Deng2009}
Deng, J., Dong, W., Socher, R., Li, L., Li, K., Li, F.: {ImageNet: {A}
  large-scale hierarchical image database}. In: 2009 {IEEE} Computer Society
  Conference on Computer Vision and Pattern Recognition. pp. 248--255. {IEEE}
  Computer Society (2009). \doi{10.1109/CVPR.2009.5206848}

\bibitem{engstrom19a}
Engstrom, L., Tran, B., Tsipras, D., Schmidt, L., Madry, A.: Exploring the
  landscape of spatial robustness. In: Proceedings of the 36th International
  Conference on Machine Learning. vol.~97, pp. 1802--1811. PMLR (2019)

\bibitem{fawzi18geometry}
Fawzi, A., Moosavi-Dezfooli, S.M., Frossard, P., Soatto, S.: Empirical study of
  the topology and geometry of deep networks. In: 2018 IEEE/CVF Conference on
  Computer Vision and Pattern Recognition. pp. 3762--3770 (2018).
  \doi{10.1109/CVPR.2018.00396}

\bibitem{Gehr2018}
Gehr, T., Mirman, M., Drachsler{-}Cohen, D., Tsankov, P., Chaudhuri, S.,
  Vechev, M.T.: {{AI2:} Safety and Robustness Certification of Neural Networks
  with Abstract Interpretation}. In: 2018 {IEEE} Symposium on Security and
  Privacy. pp. 3--18. {IEEE} Computer Society (2018).
  \doi{10.1109/SP.2018.00058}

\bibitem{Geirhos2020a}
Geirhos, R., Jacobsen, J.H., Michaelis, C., Zemel, R., Brendel, W., Bethge, M.,
  Wichmann, F.A.: Shortcut learning in deep neural networks. Nature Machine
  Intelligence  \textbf{2},  665--673 (Nov 2020)

\bibitem{Houben2013}
Houben, S., Stallkamp, J., Salmen, J., Schlipsing, M., Igel, C.: {Detection of
  traffic signs in real-world images: The German traffic sign detection
  benchmark}. In: The 2013 International Joint Conference on Neural Networks.
  pp.~1--8. {IEEE} (2013). \doi{10.1109/IJCNN.2013.6706807}

\bibitem{Huang2017}
Huang, X., Kwiatkowska, M., Wang, S., Wu, M.: {Safety Verification of Deep
  Neural Networks}. In: Majumdar, R., Kuncak, V. (eds.) Computer Aided
  Verification - 29th International Conference. Lecture Notes in Computer
  Science, vol. 10426, pp. 3--29. Springer (2017).
  \doi{10.1007/978-3-319-63387-9\_1}

\bibitem{Katz2017}
Katz, G., Barrett, C.W., Dill, D.L., Julian, K., Kochenderfer, M.J.: {Reluplex:
  An Efficient {SMT} Solver for Verifying Deep Neural Networks}. In: Majumdar,
  R., Kuncak, V. (eds.) Computer Aided Verification - 29th International
  Conference. Lecture Notes in Computer Science, vol. 10426, pp. 97--117.
  Springer (2017). \doi{10.1007/978-3-319-63387-9\_5}

\bibitem{Kim2021}
Kim, Y., Hwang, H., Shin, J.: Robust object detection under harsh
  autonomous-driving environments. IET Image Processing  (2021).
  \doi{10.1049/ipr2.12159}

\bibitem{Michaelis2019}
Michaelis, C., Mitzkus, B., Geirhos, R., Rusak, E., Bringmann, O., Ecker, A.S.,
  Bethge, M., Brendel, W.: Benchmarking robustness in object detection:
  Autonomous driving when winter is coming. CoRR  \textbf{abs/1907.07484}
  (2019)

\bibitem{Ponn2020}
Ponn, T., Kr{\"{o}}ger, T., Diermeyer, F.: Identification and explanation of
  challenging conditions for camera-based object detection of automated
  vehicles. Sensors  \textbf{20}(13), ~3699 (2020). \doi{10.3390/s20133699}

\bibitem{Singh2019}
Singh, G., Gehr, T., P{\"{u}}schel, M., Vechev, M.T.: {An abstract domain for
  certifying neural networks}. Proc. {ACM} Program. Lang.  \textbf{3}({POPL}),
  41:1--41:30 (2019). \doi{10.1145/3290354}

\bibitem{Szegedy2016}
Szegedy, C., Vanhoucke, V., Ioffe, S., Shlens, J., Wojna, Z.: {Rethinking the
  Inception Architecture for Computer Vision}. In: 2016 {IEEE} Conference on
  Computer Vision and Pattern Recognition. pp. 2818--2826. {IEEE} Computer
  Society (2016). \doi{10.1109/CVPR.2016.308}

\bibitem{Szegedy2014}
Szegedy, C., Zaremba, W., Sutskever, I., Bruna, J., Erhan, D., Goodfellow,
  I.J., Fergus, R.: Intriguing properties of neural networks. In: Bengio, Y.,
  LeCun, Y. (eds.) 2nd International Conference on Learning Representations
  (2014), \url{http://arxiv.org/abs/1312.6199}

\bibitem{temel2019traffic}
Temel, D., Chen, M., AlRegib, G.: Traffic sign detection under challenging
  conditions: A deeper look into performance variations and spectral
  characteristics. IEEE Transactions on Intelligent Transportation Systems pp.
  1--11 (2019). \doi{10.1109/TITS.2019.2931429}

\bibitem{Temel2017_NIPSW}
Temel, D., Kwon, G., Prabhushankar, M., AlRegib, G.: {CURE-TSR: Challenging
  unreal and real environments for traffic sign recognition}. In: Neural
  Information Processing Systems (NeurIPS) Workshop on Machine Learning for
  Intelligent Transportation Systems (2017)

\bibitem{adaptive20tramer}
Tramer, F., Carlini, N., Brendel, W., Madry, A.: On adaptive attacks to
  adversarial example defenses. In: Advances in Neural Information Processing
  Systems. vol.~33, pp. 1633--1645. Curran Associates, Inc. (2020)

\end{thebibliography}

\end{document}